\documentclass[conference]{IEEEtran}
\IEEEoverridecommandlockouts
\usepackage{cite}
\usepackage{amsmath,amssymb,amsfonts}
\usepackage{algorithmic}
\usepackage{graphicx}
\usepackage{textcomp}
\usepackage{xcolor}
\usepackage{multirow}
\usepackage{tabularx}
\usepackage{booktabs}
\usepackage{pifont}
\usepackage{bbding}
\usepackage{caption}
\usepackage{url}
\usepackage[colorlinks=true, 
            linkcolor=blue,       
            citecolor=green,      
            urlcolor=magenta      
           ]{hyperref}

\def\BibTeX{{\rm B\kern-.05em{\sc i\kern-.025em b}\kern-.08em
    T\kern-.1667em\lower.7ex\hbox{E}\kern-.125emX}}
\begin{document}

\title{SingingBot: An Avatar-Driven System for Robotic Face Singing Performance}

\author{Zhuoxiong Xu, Xuanchen Li, Yuhao Cheng, Fei Xu, Yichao Yan, Xiaokang Yang}

\twocolumn[{
\renewcommand\twocolumn[1][]{#1}
\maketitle
\begin{center}
    \captionsetup{type=figure}
    \includegraphics[width=\textwidth]{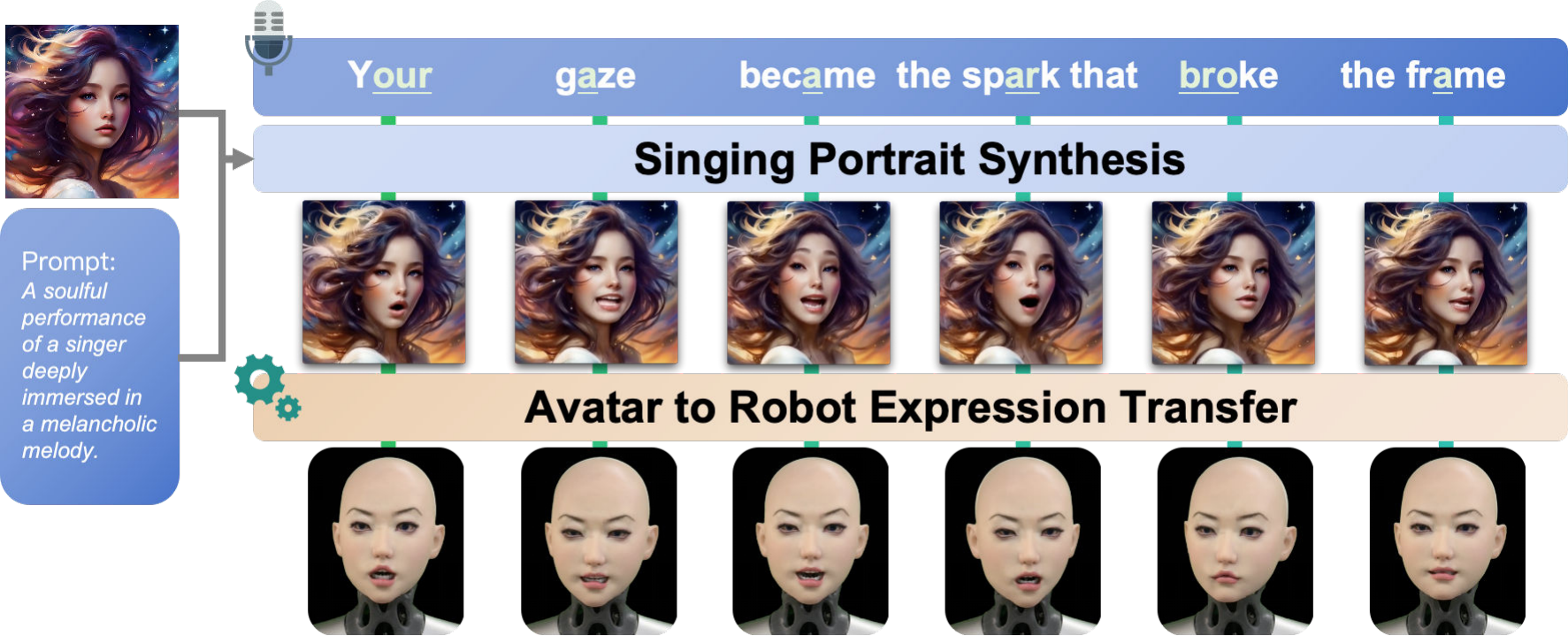}

    \captionof{figure}{We present \textbf{SingingBot}, a framework to generate robotic face singing performance from synthesized portrait animations. Our method conveys the rich emotions of singing while maintaining obvious lip-audio synchronization.}

\end{center}
}]

\begin{abstract}
Equipping robotic faces with singing capabilities is crucial for empathetic Human-Robot Interaction. However, existing robotic face driving research primarily focuses on conversations or mimicking static expressions, struggling to meet the high demands for continuous emotional expression and coherence in singing. To address this, we propose a novel avatar-driven framework for appealing robotic singing. We first leverage portrait video generation models embedded with extensive human priors to synthesize vivid singing avatars, providing reliable expression and emotion guidance. Subsequently, these facial features are transferred to the robot via semantic-oriented mapping functions that span a wide expression space. Furthermore, to quantitatively evaluate the emotional richness of robotic singing, we propose the Emotion Dynamic Range metric to measure the emotional breadth within the Valence-Arousal space, revealing that a broad emotional spectrum is crucial for appealing performances. Comprehensive experiments prove that our method achieves rich emotional expressions while maintaining lip-audio synchronization, significantly outperforming existing approaches. Project page:
\href{https://droshown.github.io/SingingBot/}{https://droshown.github.io/SingingBot/}
\end{abstract}

\begin{IEEEkeywords}
Robotic Face, Facial Animation, Motor Control
\end{IEEEkeywords}

\section{Introduction}
\label{sec:intro}


Singing is a universal way to express emotion that goes beyond language or culture. Unlike speech, singing demands continuous articulation constrained by melody and rhythm, requiring longer vowel sounds and specific lyrical expressions. Consequently, endowing humanoid robots with singing capabilities serves as a critical benchmark for evaluating human behaviour replication and represents a crucial step toward natural human-robot interaction in companionship, reception, and entertainment. However, the higher requirements for expressive coherence and emotional breadth make this task highly challenging.

Early approaches~\cite{lin2016expressional, asheber2016humanoid} to robotic facial animation primarily relied on interpolating between preset expression sets or pre-programmed hardwares~\cite{ren2016automatic}. These methods are constrained by the expression library size and struggle to capture the nuanced expressions and emotions inherent in singing. 
To address these problems, recent data-driven studies~\cite{zhu2025awakening, li2025x2c, hu2024human, zhang2025exface} focus on automatic motor control via learning from paired data. One category~\cite{li2024driving, ugotme} generates audio-sync animations using kinematics but primarily targets talking, failing to produce the lyrical singing animation. Another category~\cite{chen2021smile, liu2024unlocking} transfers image-space human expressions to robots, but typically focuses on static expression matching, neglecting the coherent emotional expression critical for singing performances. Both of these types struggle to be applied to robot singing trivially.

Distinct from previous works, we present a framework that leverages the robust human priors within video diffusion models to enable realistic robotic singing. Specifically, we employ a human-centric video diffusion transformer~\cite{hallo3} to synthesize controllable 2D portrait animations, serving as the driving source for robotic performance. We then transfer the avatar's facial features to the physical robot by  applying a semantic-oriented piecewise mapping strategy to generate motor control values. Contrary to the existing data-driven methods~\cite{zhu2025awakening, li2025x2c, hu2024human, zhang2025exface}, the semantic-based mapping spans a larger expression space, ensuring the final performance preserves both emotional nuance and lip-sync accuracy.

Furthermore, quantifying robot singing performance at the perceptual level remains a significant challenge. Existing audio-driven animation methods focus on lip synchronization but overlook emotional richness. Drawing inspiration from affective computing, we leverage the Valence-Arousal (VA) space derived from the Circumplex Model of Affect~\cite{russell1980circumplex} to capture subtle emotional differences, where any emotion can be represented by two continuous, orthogonal dimensions. Specifically, we propose the Emotion Dynamic Range (EDR) as a metric to quantify the emotional breadth. By comparing the area of the convex hull formed by the emotion trajectory in the VA space, we demonstrate that our method achieves expressive performances while maintaining precise lip synchronization, significantly outperforming the baselines.

Our contributions are summarized as follows: (1) We propose a framework to transfer prompt-controlled avatar animations to robotic faces, bridging the gap between digital humans and physical robots with vivid lip movements and emotions. 
(2) We reveal the importance of evaluating emotional richness in singing performances and propose Emotion Dynamic Range (EDR) as a quantitative indicator of emotional breadth based on the VA space.
(3) We conduct extensive experiments comparing various singing strategies, quantitatively and qualitatively demonstrating that our system achieves SOTA performance.

\section{Related Works}
\label{sec:related_works}
\textbf{Animatronic Robotic Face Control.} Generating robotic expressions is hindered by the structural gap between biological muscles and sparse motors, alongside the complex non-linear deformations of silicone skin~\cite{sheng2025review, pang2021review, faraj2021facially}.
Traditional methods~\cite{fukuda2002facial, lin2016expressional, asheber2016humanoid} rely on interpolating predefined expression bases but lack expressiveness and generalization. Recent learning-based approaches~\cite{zhu2025awakening, zhang2025exface} utilize neural networks to map audio or images to motor parameters either directly or via a trained imitator~\cite{chen2021smile, zhang2025morpheus}. However, constrained by the scarcity of paired data, these works typically target simple conversational scenarios, often neglecting temporal coherence and continuous emotional expression. While works like UGotMe~\cite{ugotme} introduce emotion classification for empathetic interaction, they rely on fixed emotion bases and fail to jointly generate emotional expressions with synchronized lip movements for singing. Our work distinguishes itself by focusing on robotic singing, utilizing large-scale human priors to achieve high realism and emotional richness.

\textbf{Audio Driven Avatar Facial Animation.} A fundamental challenge in audio-driven avatar and robot animation lies in modeling the complex mapping between audio and facial dynamics. The 3D talking head methods~\cite{emotalk,chung2025audio2face} often struggle to generalize to exaggerated expressions or singing due to limited annotated data. Conversely, 2D methods~\cite{hallo3,zhang2023sadtalker} leverage web-scale data to extract robust human priors, enabling the generation of realistic videos with rich emotions, micro-expressions, and prompt-based control. We leverage these pre-trained 2D human priors as a high-fidelity and flexible driving source, bridging the gap between avatars and robots. By transferring these virtual avatar facial features to the physical robotic domain, we endow the robot with singing capabilities, satisfying the high requirements for continuous emotional and expressive articulation.
\section{Methodology}
\label{sec:method}
\begin{figure*}[t]
  \centering
  \includegraphics[width=\linewidth]{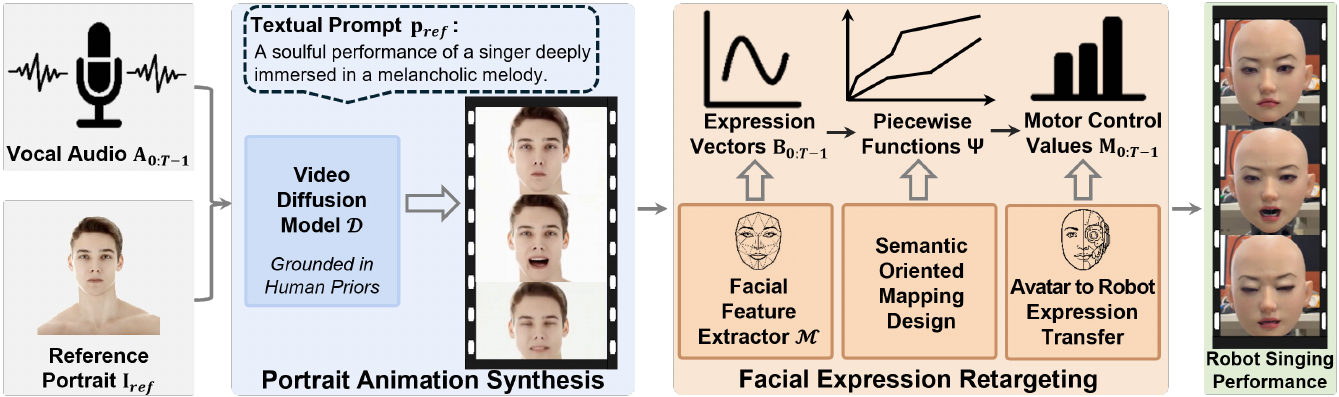}
  \caption{\textbf{The overall pipeline of SingingBot for robotic singing performance.} Given a vocal audio and a reference portrait, our method first synthesizes a vivid avatar singing animation using a pretrained video diffusion model. Benefiting from the embedded extensive expression and emotion priors, the avatar animation serves as a reliable driving source that provides expressive facial features for subsequent robotic performance. Through the semantic-oriented piecewise functions, the avatar's facial features are then mapped to the physical robot motion space, achieving coherent and appealing robotic singing performance.}
  \label{fig:pipeline}
\end{figure*}
Our goal is to generate lyrical robotic facial animation synchronized with the vocals of a song. Specifically, given $\mathbf{T}$ frame vocal $\mathbf{A}_{0:T-1}$, our framework generates robotic facial motions $\mathbf{M}_{0:T-1}=\{\mathbf{m}_i\in \mathbb{R}^{d}\}_{i=0}^{T-1}$ with each motion vector having $d$ Degrees of Freedom (DoFs) represented by a control value ranging from 0 to 1.

To this end, we propose \textbf{SingingBot}, as illustrated in Fig.~\ref{fig:pipeline}. First, we employ a pretrained video diffusion model, grounded in extensive human priors, to synthesize realistic 2D portrait videos from vocal audio (Sec.~\ref{sec:animation_synthesis}). These generated videos serve as the intermediate driving source for our robotic face. Our method then utilizes a facial feature extractor to derive expression vectors from the synthesized video. Subsequently, we apply a well-designed piecewise mapping function to translate these expression vectors into motor actuation parameters, yielding the final robotic facial animation (Sec.~\ref{sec:exp_mapping}).

\subsection{Portrait Animation Synthesis} 
\label{sec:animation_synthesis}
Directly regressing motor parameters poses a significant challenge due to the scarcity of large-scale paired motor and expression data. Consequently, many prior works leverage intermediate representations, such as facial landmarks or drivable robot imitators, to drive robotic faces. However, these approaches struggle to be applied to singing scenarios. Since singing performances demand continuous and emotionally charged expressions, landmark-driven methods struggle to convey emotion due to their sparsity and lack of appearance. Similarly, virtual robot imitators, constrained by limited training data, fail to generalize to the diverse emotions and complex expressions required for realistic singing. Inspired by the recent progress of video diffusion transformers pretrained on large-scale human-centric data, we posit that the embedded extensive human priors can provide sufficient facial expression and emotional guidance for robotic face singing animations. 
Specifically, we employ a pre-trained video diffusion model $\mathcal{D}$ to synthesize 2D singing videos $\mathbf{V}_{0:T-1}=\{\mathbf{v}_i\in \mathbb{R}^{h \times w \times 3}\}_{i=0}^{T-1}$ aligned with the input vocal. To control the style of the singing performance, we utilize an reference portrait image $\mathbf{I}_{ref}$ and a text prompt $\mathbf{p}_{ref}$ as conditions:
\begin{equation}
    \mathbf{V}_{0:T-1}=\mathcal{D}(\mathbf{A}_{0:T-1}, \mathbf{I}_{ref}, \mathbf{p}_{ref})
    \label{eq:diffusion}.
\end{equation}
The extensive human priors ensure that the synthesized singing videos exhibit rich emotional fidelity and coherent expressions. 

\subsection{Facial Expression Retargeting}
\label{sec:exp_mapping}
Leveraging the synthesized portrait singing video as the driving source, we proceed to retarget the expressive facial dynamics onto the robot's physical action space.

\textbf{Avatar Facial Dynamics Extraction.}
We adopt the 52-dimensional blendshape coefficient vector that complies with the ARKit standard~\cite{arkit} as the parameterized representation of facial expressions. These coefficients quantify the intensity of specific muscle actions, e.g., \textit{jawOpen} and \textit{mouthStretch}. Unlike previous methods relying on raw 2D facial landmarks, blendshapes offer a semantically meaningful feature space that is identity-invariant and highly compatible with existing animation pipelines. Specifically, we employ MediaPipe $\mathcal{M}$~\cite{mediapipe} to extract frame-wise blendshape coefficients $\mathbf{B}_{0:T-1}=\{\mathbf{b}_i\in \mathbb{R}^{52}\}_{i=0}^{T-1}$ from the synthesized video:
\begin{equation}
    \mathbf{B}_{0:T-1} = \mathcal{M}(\mathbf{V}_{0:T-1}).
    \label{eq:bs_extraction}
\end{equation}

We post-process the extracted expression vectors using Gaussian smoothing to alleviate the temporal jitter that can cause jerky robotic motion. This ensures smooth motion dynamics while preserving significant expressive details.

\textbf{Semantic-Oriented Avatar to Robot Expression Transfer.}
The core challenge in driving the robot face lies in mapping visual expression features to the robotic motion space. To circumvent the generalization limitations inherent in previous data-driven approaches, we avoid learning black-box mappings from paired datasets. Instead, we propose a \textbf{Semantic-Oriented Piecewise Function Design}, which maps blendshape parameters to motor control values based on semantic correspondence. Formally, given the intensity scalar $\beta_{j} \in \mathbf{b}$ of the $j$-th ARKit-compliant semantic expression, we define its corresponding function $\Psi_j(\cdot)$ as:
\begin{equation}
    \Delta{\mathbf{m}}_j = \Psi_j(\beta_j) = \mathbf{w}_{j,k} \cdot \beta_j + \mathbf{c}_{j,k}, \quad \text{for } \beta_j \in [\tau_{j,k}, \tau_{j,k+1}),
    \label{eq:mapping_def}
\end{equation}
where $\Delta\mathbf{m}_j \in \mathbb{R}^{d}$ represents the contributing motion control value to the robot's $d$ actuators. The function $\Psi_j$ is defined by a set of $K$ intervals, where $\mathbf{w}_{j,k}$ and $\mathbf{c}_{j,k}$ represent the slope and intercept vectors for the $k$-th linear segment bounded by thresholds $[\tau_{j,k}, \tau_{j,k+1})$. We manually design these piecewise functions for each semantic expression, and highly non-linear expression bases typically require denser pieces to ensure expressiveness.

Since the DoF of the robot is significantly sparser than the blendshapes, certain semantic expressions (e.g., \textit{cheekPuff}) lack direct physical counterparts and are excluded as $\Psi_j(\beta_j)=0$. Furthermore, to address mechanical constraints where some of the asymmetric expressions (e.g., \textit{noseSneerLeft/Right}) share a single actuator, we unify them into a single symmetric mapping and use their average as the input density. Please refer to the Supplementary Material for design details. 

Finally, by blending the control values obtained from all valid mappings, we obtain the final control value $\mathbf{m}$:
\begin{equation}
    \mathbf{m} = \mathbf{m}_{rest} + \sum\nolimits_{{\beta}_j \in \mathbf{b}} \Psi_j({\beta}_j),
    \label{eq:final_motor}
\end{equation}
where $\mathbf{m}_{rest}$ denotes the initial rest pose.
Additionally, we linearly map the 3-DoF pose extracted via MediaPipe~\cite{mediapipe} to the three neck motors to control head motions.
\section{EXPERIMENTS}
\label{sec:eval}
\subsection{Implementation Details}
\textbf{Robotic Face Platform.}
We test our method on the \textbf{Hobbs} robot platform, which features 32 Degrees of Freedom, including 29 motors for facial motions and 3 for head/neck movements. The motors are connected to silicone skin through articulated linkages. The portrait animations are synthesized using Hallo3~\cite{hallo3} on a remote server equipped with a single NVIDIA A800 GPU, while the expression transfer module is executed locally on the robot's embedded RK3588 processor.

\textbf{Data Preparation.} We collected a test set comprising 40 singing vocal clips, each ranging from 3 to 4 seconds in duration. The dataset encompasses multiple languages and emotional styles to ensure diversity in phonetic coverage and emotional expression.

\textbf{Baselines.}
We evaluate our method against three data-driven baselines: \textbf{(1)} Random Sampling (RT), which randomly samples control values from the training set; \textbf{(2)} Nearest Neighbor Retrieval~(NNR), which finds the closest sample in the training set based on blendshapes; and \textbf{(3)} a direct regression approach proposed by Zhu~et~al.~\cite{zhu2025awakening}, which maps blendshape coefficients directly to motor control values. We generate blendshapes using EmoTalk~\cite{emotalk}. 
All baselines are re-implemented and trained on our collected dataset, consisting of 10K randomly sampled pairs.
 
\subsection{Quantitative Comparisons}
We utilize the widely used Lip Sync Error Distance (LSE-D) and Lip Sync Error Confidence (LSE-C) to measure the synchronization between the singing audio and the robotic performance~\cite{w2l}. A lower LSE-D value indicates a higher consistency between the lip and the audio, while a higher LSE-C reflects a  better alignment between the audio and the performance. Additionally, we propose the Emotion Dynamic Range~(EDR) metric to quantify the emotional breadth based on the Valence-Arousal~(VA) space. Specifically, we utilize a pre-trained emotion recognition model $\Phi(\cdot)$~\cite{savchenko2022classifying, savchenko2023facial} to extract the per-frame emotion coordinates from the robotic performance, yielding a set of discrete VA-space points $\mathcal{P} = \{p_t \mid p_t = (v_t, a_t)\}_{t=0}^{T-1}$, where $v_t, a_t \in [-1, 1]$ represent valence and arousal. Formally, let $\mathcal{H}$ be the smallest convex polygon containing all points in $\mathcal{P}$ after removing the top $5\%$ of outliers. The EDR is calculated as the geometric area of $\mathcal{H}$. A larger EDR value indicates a wider distribution of emotional states, reflecting a more expressive performance.

The quantitative results are reported in Tab.~\ref{tab:metric}. Our method significantly outperforms the baselines in terms of both lip-audio synchronization and emotional expressiveness. Notably, our method achieves an EDR one order of magnitude higher than the competing methods. While RT covers a broad expression space, it tends to generate unnatural expressions misclassified as ``\textit{Surprise}," whereas NNR and Zhu~et~al.~\cite{zhu2025awakening} fail to produce rich emotions or expressive lip movements. Our method benefits from human priors and semantic-oriented mapping, allowing it to exhibit rich emotions while maintaining high-quality lip-sync. Besides, by utilizing a photorealistic avatar to provide reliable expression and emotion signals, our method yields significantly enhanced generalizability and emotion diversity compared with the data-limited methods. 

\begin{figure}[t]
  \centering
  \includegraphics[width=\linewidth]{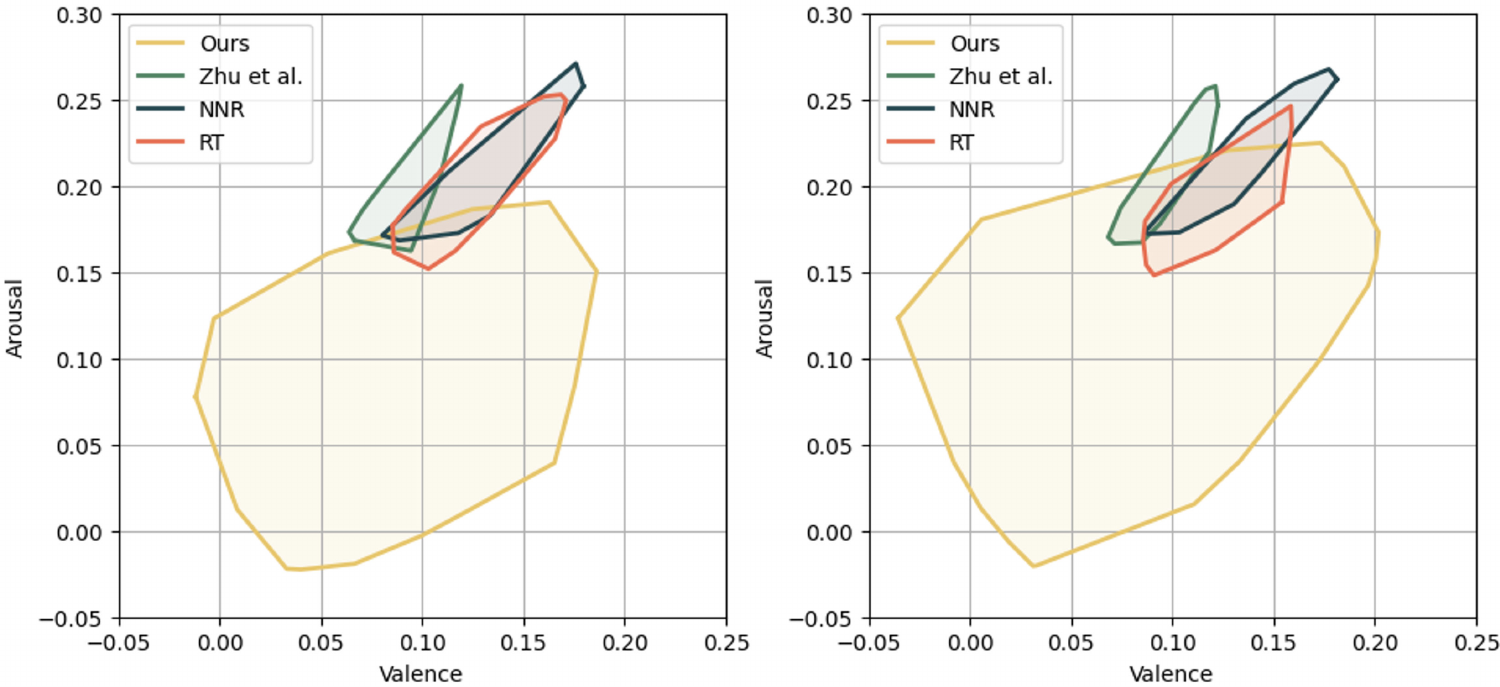}
  \caption{\textbf{Visualization of Emotion Dynamic Range.} The polygonal area represents the emotional richness of singing performance. Our method achieves significantly better emotional expressiveness.}
  \label{fig:edr}
\end{figure}
\textbf{Visualization of Emotion Dynamic Range.} The Emotion Dynamic Range~(EDR) metric reflects the distribution range of emotions within singing performances in the Valence-Arousal (VA) space. To intuitively visualize the geometric interpretation of EDR, we sample two clips and plot the results. As shown in the Fig.~\ref{fig:edr}, the emotional ranges of different methods are depicted by polygons of varying colors, where the polygonal area stands for the EDR metric.
Since NNR and Zhu~et~al.~\cite{zhu2025awakening} generate ambiguous lip motions without distinct emotional state, they occupy only a small region. Although RT covers a vast area of robotic expression space, the generated expressions typically lack valid emotional semantics, resulting in a limited EDR. In contrast, our method demonstrates significantly richer emotional expressiveness, which is critical for achieving engaging singing performances.


\begin{table}[t]
\caption{Quantitative comparisons with baseline methods in terms of facial expressions and emotion richness.}
\resizebox{\columnwidth}{!}{%
\begin{tabular}{llll}
\toprule
Method     & LSE-D~$\downarrow$ & LSE-C~$\uparrow$ & EDR~$\uparrow$ \\ \hline
RT   &  14.892 &  0.196 &  0.0044     \\
NNR   &   12.834   &   1.481   &  0.0030            \\
Zhu et al.~\cite{zhu2025awakening} &   12.428   &  1.504    & 0.0021        \\
Ours       &   \textbf{11.095}   &   \textbf{2.313}   &  \textbf{0.0389}      \\ \bottomrule
\end{tabular}%
}
\label{tab:metric}
\end{table}
\begin{table}[t]
\caption{User study. We evaluate audience preferences in terms of realism, emotional resonance, and lip-sync.}
\resizebox{\columnwidth}{!}{%
\begin{tabular}{llll}
\toprule
Method     & Realism~$\uparrow$ & Resonance~$\uparrow$ & Lip-Sync~$\uparrow$ \\ \hline
RT   & 1.21 & 1.32  &   1.33       \\
NNR   & 1.50  & 1.57  &   1.87       \\
Zhu et al.~\cite{zhu2025awakening}  &  1.94  &    1.73  &     2.21     \\
Ours    &   \textbf{3.02}  &  \textbf{3.05}  &    \textbf{3.57}   \\ \bottomrule
\end{tabular}%
}
\label{tab:user}
\end{table}
\subsection{User Study}
We conduct a user study to further evaluate the audience's preferences for robotic singing performances generated by different methods. Specifically, the study consists of 16 questions, each presenting the participants with a 7-second-long clip
of the singing performance. Participants are asked to rate on a scale
of 1 to 5 according to the expression realism, emotional resonance, and lip-audio synchronization of the performance. 
We eventually receive 40 valid submissions.
As is demonstrated in Tab.~\ref{tab:user}, our method
produces the most realistic singing performance and excels
at emotional expressiveness, achieving the
best perceptual quality.

\subsection{Qualitative Comparisons}
\begin{figure}[t]
  \centering
  \includegraphics[width=\linewidth]{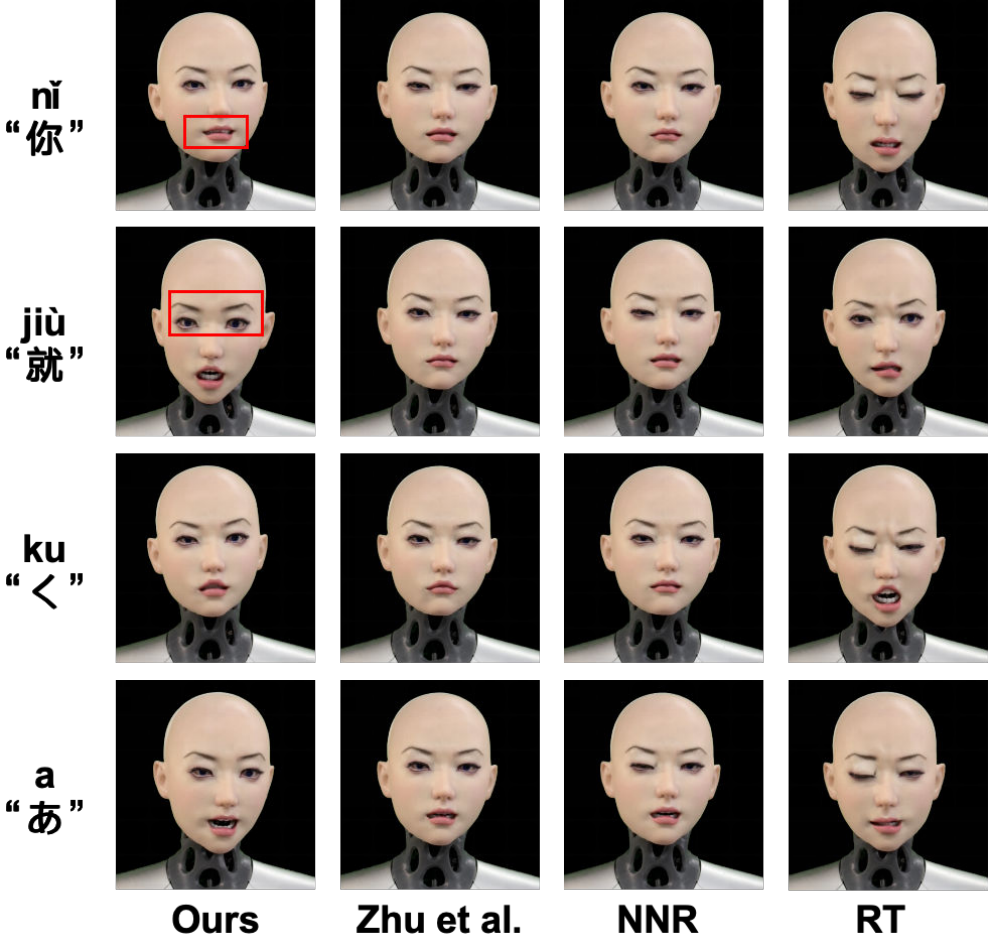}
  \caption{\textbf{Visual comparison of generated robotic singing performances.} Our method displays rich emotions while presenting plausible lip shapes. The red box highlights the micro expressions expressing subtle emotions.}
  \label{fig:baseline}
\end{figure} Fig.~\ref{fig:baseline} shows a visual comparison between the robotic performance generated by different methods. RT uses random sampling from the dataset, resulting in intense but unnatural expressions that produce the Uncanny Valley effect. Although NNR and Zhu~et~al.~\cite{zhu2025awakening} generate lip movements that are somewhat consistent with the audio, the results are overly conservative and lack expressiveness. In contrast, our method produces more plausible lip shapes, effectively capturing the large-amplitude movements and prolonged vowels featured in singing performances. Furthermore, while maintaining lip synchronization, our method simultaneously exhibits micro-expressions that convey rich emotions, such as the slightly raised mouth corners (Row 1) and widened eyes (Row 2).


\subsection{Ablation Study}
\begin{table}[t]
\caption{Ablation study of our method in terms of the influence of different driving strategies.}
\resizebox{\columnwidth}{!}{%
\begin{tabular}{llll}
\toprule
Method                & LSE-D~$\downarrow$ & LSE-C~$\uparrow$  & EDR~$\uparrow$ \\ \hline

w/o Avatar-driven     &        13.069  & 1.023   & 0.0038\\
w/o Diffusion Prior   &        14.387  & 1.048    & 0.0035 \\
Ours                  &     \textbf{11.095}   &   \textbf{2.313} & \textbf{0.0389}  \\ \bottomrule
\end{tabular}%
}
\label{tab:ablation}
\end{table}
\begin{figure}[t]
  \centering
  \includegraphics[width=\linewidth]{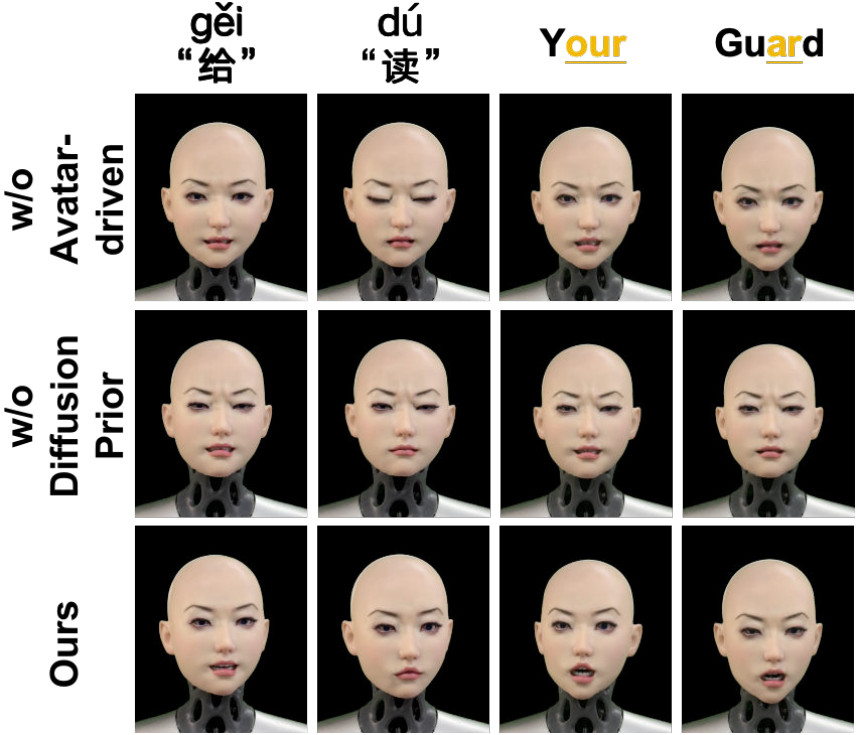}
  \caption{\textbf{Visual comparison of different singing strategies.} Using avatar videos embedded with diffusion priors as driving sources significantly enhances the animation performance of robot singing.}
  \label{fig:ablation}
\end{figure}
\textbf{Analysis of Avatar-based Driving.} To assess the contribution of the avatar-based driving source, we excluded the intermediate avatar video generation phase. Instead, we employ the blendshape coefficients predicted by EmoTalk~\cite{emotalk} as the input to our piecewise functions (denoted as \textbf{w/o Avatar-driven}). As shown in Tab.~\ref{fig:ablation}, all metrics significantly decrease. The results in Fig.~\ref{fig:ablation} show that the robot can only produce ambiguous mouth shapes and slight emotional changes. In contrast, our method can transfer the visual expression and emotion features embedded in the portrait animation, achieving expressive performance.

\textbf{Analysis of Diffusion Human Priors.} Video diffusion models leverage extensive human priors to achieve portrait animation quality significantly superior to methods trained on limited datasets. To investigate the impact of diffusion priors on robotic animation quality, we utilize rendered avatar animations generated by EmoTalk~\cite{emotalk} as the driving animation (denoted as \textbf{w/o Diffusion Prior}). All metrics decline in this setting, as reported in Tab.~\ref{tab:ablation}. This degradation demonstrates that although rendered 3D avatar animations provide basic visual motion signals, they lack the subtle micro-expressions and appearance features embedded in diffusion priors. These features are crucial for recognizing emotional state and expression strength. Consequently, the resulting robotic performance suffers from constrained lip movements and emotional misalignment, as illustrated in Fig.~\ref{fig:ablation}.

\textbf{Analysis of Reference Portrait.} 
\begin{figure}[t]
  \centering
  \includegraphics[width=\linewidth]{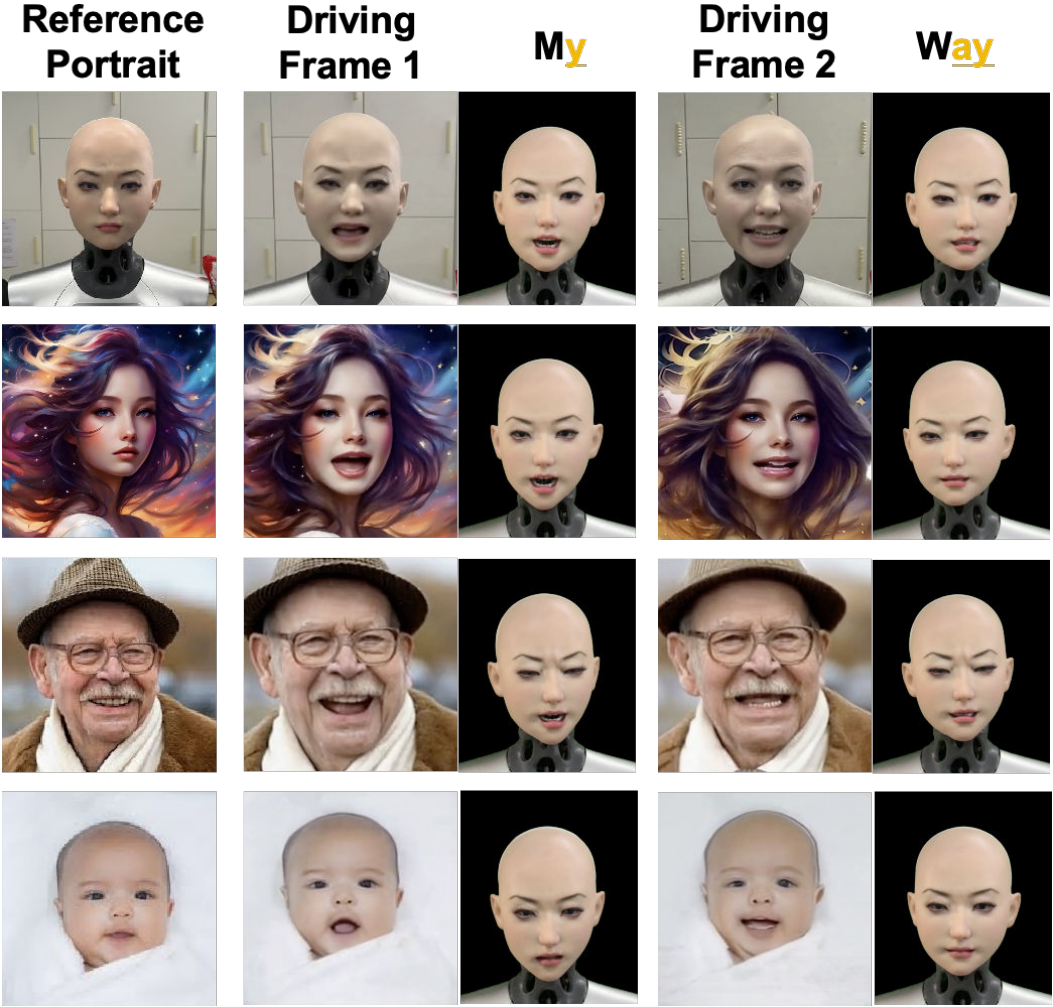}
  \caption{\textbf{Singing style control.} Our approach achieves stylistic  performances driven by different reference portraits.}
  \label{fig:style}
\end{figure}
Our method can benefit from the controllability of the diffusion generation model, generating different styles of singing performances by changing conditions, and further achieving diverse robotic face animations.
Fig.~\ref{fig:style} showcases the robotic face performance generated from the same vocal input, using 4 distinct reference portraits to synthesize the driving animation. By switching reference portraits, our method enables diverse singing performance styles.
\section{Conclusion}
\label{sec:conclusion}
In this paper, we present SingingBot, a novel framework for robotic singing performance. 
To address the limitations of existing methods in handling singing scenarios which demand superior lip articulation and emotional expressiveness, our method utilizes a portrait animation model pre-trained on large-scale data to generate realistic avatar singing videos as driving sources, leveraging embedded human priors to provide abundant expression and emotion guidance.
Through a set of semantic-oriented mapping functions, our method transfers virtual facial features to the physical robotic motion space, achieving appealing robotic singing performances.
Furthermore, we introduce the Emotion Dynamic Range metric to quantify emotional diversity. By analyzing the emotional breadth within the Valence-Arousal space, we reveal that a rich emotional spectrum is critical for performance expressiveness.
Extensive experiments prove that our method achieves superior performance over existing methods both in lip-audio synchronization and emotional richness, taking a meaningful step towards empathetic Human-Robot Interaction. 

\bibliographystyle{IEEEbib}
\bibliography{icme2025references}

@inproceedings{zhu2025awakening,
  title={Awakening Facial Emotional Expressions in Human-Robot},
  author={Zhu, Yongtong and Li, Lei and Qian, Iggy and Zhou, WenBin and Yuan, Ye and Li, Qingdu and Liu, Na and Zhang, Jianwei},
  booktitle={IROS},
  pages={21511--21518},
  year={2025}
}

@inproceedings{fukuda2002facial,
  title={Facial expression of robot face for human-robot mutual communication},
  author={Fukuda, Toshio and Taguri, Jun and Arai, Fumihito and Nakashima, Masakazu and Tachibana, Daisuke and Hasegawa, Yasuhisa},
  booktitle={ICRA},
  year={2002}
}

@article{pang2021review,
  title={Review of robot skin: A potential enabler for safe collaboration, immersive teleoperation, and affective interaction of future collaborative robots},
  author={Pang, Gaoyang and Yang, Geng and Pang, Zhibo},
  journal={IEEE Transactions on Medical Robotics and Bionics},
  pages={681--700},
  year={2021}
}

@article{sheng2025review,
  title={A review on an AI-driven face robot for human-robot expression interaction},
  author={Sheng, Qincheng and Tang, Wei and Qin, Hao and Kong, Yujie and Dai, Haokai and Zhong, Yiding and Wang, Yonghao and Zou, Jun and Yang, Huayong},
  journal={Science China Technological Sciences},
  pages={2010301},
  year={2025}
}

@article{liu2024unlocking,
  title={Unlocking human-like facial expressions in humanoid robots: A novel approach for action unit driven facial expression disentangled synthesis},
  author={Liu, Xiaofeng and Ni, Rongrong and Yang, Biao and Song, Siyang and Cangelosi, Angelo},
  journal={IEEE Transactions on Robotics},
  pages={3850--3865},
  year={2024}
}

@article{zhang2025morpheus,
  title={Morpheus: A Neural-driven Animatronic Face with Hybrid Actuation and Diverse Emotion Control},
  author={Zhang, Zongzheng and Yang, Jiawen and Peng, Ziqiao and Yang, Meng and Ma, Jianzhu and Cheng, Lin and Xu, Huazhe and Zhao, Hang and Zhao, Hao},
  journal={arXiv preprint arXiv:2507.16645},
  year={2025}
}

@inproceedings{chen2021smile,
  title={Smile like you mean it: Driving animatronic robotic face with learned models},
  author={Chen, Boyuan and Hu, Yuhang and Li, Lianfeng and Cummings, Sara and Lipson, Hod},
  booktitle={ICRA},
  year={2021}
}

@article{faraj2021facially,
  title={Facially expressive humanoid robotic face},
  author={Faraj, Zanwar and Selamet, Mert and Morales, Carlos and Torres, Patricio and Hossain, Maimuna and Chen, Boyuan and Lipson, Hod},
  year={2021}
}

@article{ren2016automatic,
  title={Automatic facial expression learning method based on humanoid robot XIN-REN},
  author={Ren, Fuji and Huang, Zhong},
  journal={IEEE Transactions on Human-Machine Systems},
  year={2016}
}

@inproceedings{li2024driving,
  title={Driving Animatronic Robot Facial Expression From Speech},
  author={Li, Boren and Li, Hang and Liu, Hangxin},
  booktitle={IROS},
  year={2024}
}

@misc{arkit,
  author = {{Apple Inc.}},
  title = {{ARKit}},
  howpublished = {\url{https://developer.apple.com/augmented-reality/arkit/}},
  year = {2025}
}

@article{chung2025audio2face,
  title={Audio2Face-3D: Audio-driven Realistic Facial Animation For Digital Avatars},
  author={Chung, Chaeyeon and Fedorov, Ilya and Huang, Michael and Karmanov, Aleksey and Korobchenko, Dmitry and Ribera, Roger and Seol, Yeongho and others},
  journal={arXiv preprint arXiv:2508.16401},
  year={2025}
}

@inproceedings{zhang2023sadtalker,
  title={Sadtalker: Learning realistic 3d motion coefficients for stylized audio-driven single image talking face animation},
  author={Zhang, Wenxuan and Cun, Xiaodong and Wang, Xuan and Zhang, Yong and Shen, Xi and Guo, Yu and Shan, Ying and Wang, Fei},
  booktitle={CVPR},
  pages={8652--8661},
  year={2023}
}

@article{zhang2025exface,
  title={ExFace: Expressive Facial Control for Humanoid Robots with Diffusion Transformers and Bootstrap Training},
  author={Zhang, Dong and Peng, Jingwei and Jiao, Yuyang and Gu, Jiayuan and Yu, Jingyi and Chen, Jiahao},
  journal={arXiv preprint arXiv:2504.14477},
  year={2025}
}

@article{hu2024human,
  title={Human-robot facial coexpression},
  author={Hu, Yuhang and Chen, Boyuan and Lin, Jiong and Wang, Yunzhe and Wang, Yingke and Mehlman, Cameron and Lipson, Hod},
  journal={Science Robotics},
  pages={eadi4724},
  year={2024}
}

@article{li2025x2c,
  title={X2C: A Dataset Featuring Nuanced Facial Expressions for Realistic Humanoid Imitation},
  author={Li, Peizhen and Cao, Longbing and Wu, Xiao-Ming and Yang, Runze and Yu, Xiaohan},
  journal={arXiv preprint arXiv:2505.11146},
  year={2025}
}

@inproceedings{w2l,
  title={A lip sync expert is all you need for speech to lip generation in the wild},
  author={Prajwal, KR and Mukhopadhyay, Rudrabha and Namboodiri, Vinay P and Jawahar, CV},
  booktitle={ACM MM},
  pages={484--492},
  year={2020}
}

@inproceedings{emotalk,
  title={Emotalk: Speech-driven emotional disentanglement for 3d face animation},
  author={Peng, Ziqiao and Wu, Haoyu and Song, Zhenbo and Xu, Hao and Zhu, Xiangyu and He, Jun and Liu, Hongyan and Fan, Zhaoxin},
  booktitle={ICCV},
  year={2023}
}

@inproceedings{ugotme,
  title={Ugotme: An embodied system for affective human-robot interaction},
  author={Li, Peizhen and Cao, Longbing and Wu, Xiao-Ming and Yu, Xiaohan and Yang, Runze},
  booktitle={ICRA},
  pages={5542--5548},
  year={2025},
  organization={IEEE}
}

@article{asheber2016humanoid,
  title={Humanoid head face mechanism with expandable facial expressions},
  author={Asheber, Wagshum Techane and Lin, Chyi-Yeu and Yen, Shih Hsiang},
  journal={International Journal of Advanced Robotic Systems},
  year={2016}
}

@article{lin2016expressional,
  title={An expressional simplified mechanism in anthropomorphic face robot design},
  author={Lin, Chyi-Yeu and Huang, Chun-Chia and Cheng, Li-Chieh},
  journal={Robotica},
  year={2016},
}

@inproceedings{hallo3,
  title={Hallo3: Highly dynamic and realistic portrait image animation with video diffusion transformer},
  author={Cui, Jiahao and Li, Hui and Zhan, Yun and Shang, Hanlin and Cheng, Kaihui and Ma, Yuqi and Mu, Shan and Zhou, Hang and Wang, Jingdong and Zhu, Siyu},
  booktitle={CVPR},
  pages={21086--21095},
  year={2025}
}

@article{savchenko2022classifying,
  title={Classifying emotions and engagement in online learning based on a single facial expression recognition neural network},
  author={Savchenko, Andrey V and Savchenko, Lyudmila V and Makarov, Ilya},
  journal={IEEE Transactions on Affective Computing},
  year={2022},

}

@article{russell1980circumplex,
  title={A circumplex model of affect.},
  author={Russell, James A},
  journal={Journal of personality and social psychology},
  year={1980},
  publisher={American Psychological Association}
}

@inproceedings{savchenko2023facial,
  title = 	 {Facial Expression Recognition with Adaptive Frame Rate based on Multiple Testing Correction},
  author =       {Savchenko, Andrey},
  booktitle = 	 {ICML},
  year = 	 {2023}
}

@article{mediapipe,
  title={Mediapipe: A framework for building perception pipelines},
  author={Lugaresi, Camillo and Tang, Jiuqiang and Nash, Hadon and McClanahan, Chris and Uboweja, Esha and Hays, Michael and Zhang, Fan and Chang, Chuo-Ling and Yong, Ming Guang and Lee, Juhyun and others},
  journal={arXiv preprint arXiv:1906.08172},
  year={2019}
}

\clearpage
\renewcommand{\theequation}{\Alph{equation}}
\renewcommand{\thefigure}{\Alph{figure}}
\renewcommand{\thetable}{\Alph{table}}
\setcounter{section}{0}
\setcounter{equation}{0}
\setcounter{figure}{0}
\setcounter{table}{0}

\def\BibTeX{{\rm B\kern-.05em{\sc i\kern-.025em b}\kern-.08em
    T\kern-.1667em\lower.7ex\hbox{E}\kern-.125emX}}

\renewcommand{\thetable}{\Alph{table}}
\renewcommand{\thefigure}{\Alph{figure}}
\renewcommand\thesection{\Alph{section}}

\title{SingingBot: An Avatar-Driven System for Robotic Face Singing Performance}
\author{Supplementary Material}

\maketitle
In the supplementary material, we provide further information about our work, including the specific design of our user study, the hardware design of our humanoid robots, and more details about our mapping function. We also present a short description of our supplementary video and mapping function configuration data. The
code and configuration data will be publicly released after publication.
\section{User Study Details}
\begin{figure}
    \centering
    \includegraphics[width=0.9\linewidth]{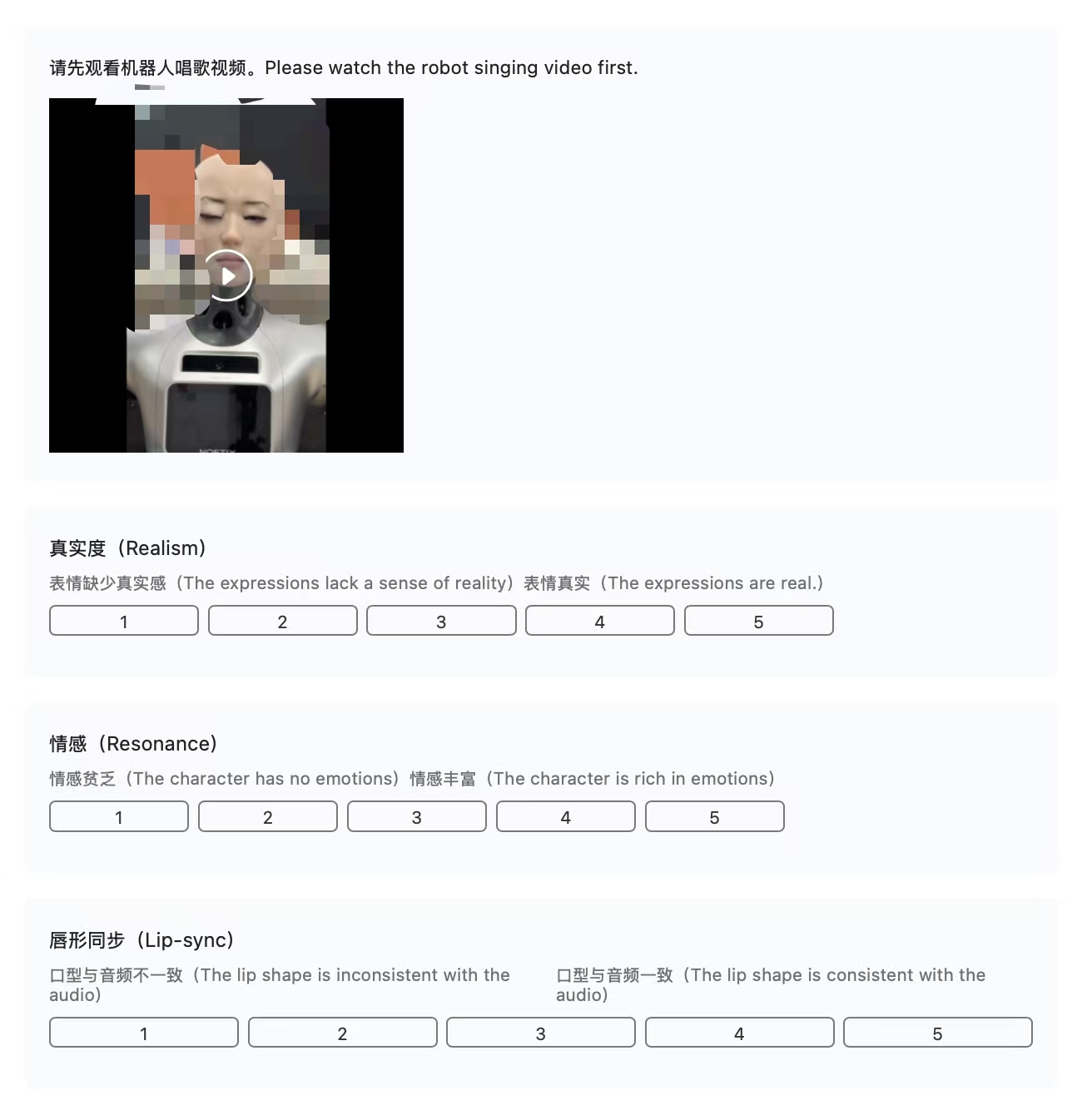}
    \caption{Demonstration of the user study interface design.}
    \label{fig:survey}
\end{figure}
Fig.~\ref{fig:survey} shows the designed user study interface. The study consists of 16 questions, each presenting the participants with a random 7-second-long clip of the robot singing performance driven by baselines and our method. The order of different methods is disrupted. For each video, participants are instructed to rate the anonymous animation on a scale of 1-5 based on the ground truth, with higher ratings indicating better performance. Each question consists of three sub-items: ``Realism: are the expressions real or are the expressions lack a sense of reality?", ``Reasonance: Is the character rich in emotions?", and ``Lip-sync: Is the lip shape consistent with the audio?" To ensure the participants are fully engaged in the user study, only when they watch the entire video and rate all the items can they proceed to the next question, otherwise a warning message would pop up. Only when all questions are answered will they be included in the final results. Finally, we collect valid submissions from 40 participants.

\section{Hardware Design}
\begin{figure}
    \centering
    \includegraphics[width=0.9\linewidth]{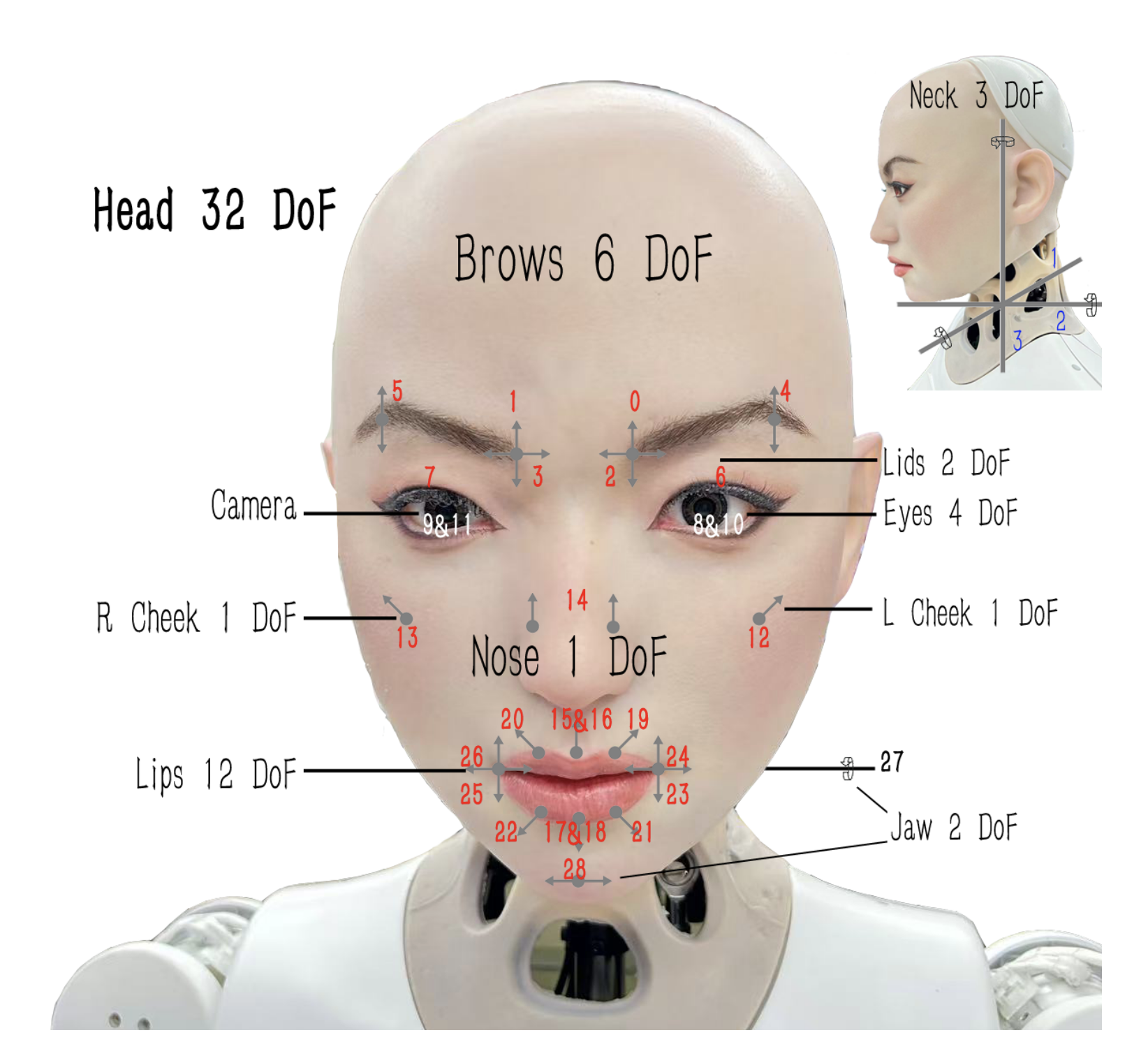}
    \caption{Facial control point degrees of freedom for our robot.}
    \label{fig:dof}
\end{figure}
To evaluate the physical performance of our SingingBot framework, we utilize a high-fidelity humanoid robotic platform as the hardware backend. The robotic head is specifically engineered for expressive facial animation, featuring a bio-inspired mechanical structure covered by a flexible silicone skin.

The robotic head integrates a total of 32 independent DoF, enabling the execution of complex micro-expressions and high-frequency lip movements required for singing. The distribution of these actuators is strategically designed to mimic human facial musculature, as illustrated in Fig.~\ref{fig:dof}.

\section{Mapping Function Details}
To bridge the gap between the generated digital facial coefficients and the physical robotic hardware, we implement a robust mapping pipeline, termed BS2Action. This process translates 52 standard ARKit-compatible Blendshapes into control signals for the 32 DoF of our robotic platform.

The mapping is grounded in a set of high-fidelity anchor pairs created by the assistance of professional animators. For each key Blendshape (e.g., JawOpen, MouthFunnel, BrowInnerUp), we chose one or more intensities, and an animator manually adjusted the robot’s 32 actuators to achieve the most precise similarity. Given the set of manually defined anchor points, we employ a piecewise interpolation strategy to determine the actuator values for any arbitrary set of Blendshape weights.

The complete configuration file and the mapping dictionary are provided in the attached supplementary ZIP archive for further reference. In the yaml file, each Blendshape has one or more servo parameter files of different Blendshape intensities, which contain the rotation angles of each servo.

\section{Video Dynamic Results}
We provide additional dynamic experimental results in the supplementary video for better demonstration of our work. Specifically, we present visual comparisons against baseline methods and ablation variants. Additionally, the video also shows the dynamic results of our method in generating different styles of performances by changing the reference portrait. Finally, we include 3 continuous singing performance lasting approximately 30 seconds to demonstrate the temporal stability of our method.

\end{document}